\documentclass[10pt,twocolumn,letterpaper]{article}

\usepackage{cvpr}              

\usepackage{graphicx}
\usepackage{amsmath}
\usepackage{amssymb}
\usepackage{booktabs}

\usepackage{color}
\usepackage{xcolor} 
\usepackage{colortbl}

\definecolor{Gray}{gray}{0.9}

\usepackage{bm}
\usepackage{bbm}
\usepackage{wrapfig}
\usepackage{amsmath}
\usepackage{graphicx}
\usepackage{amssymb}
\usepackage{booktabs}
\usepackage{multirow}
\usepackage{makecell}
\usepackage{pifont}
\usepackage{algorithm}
\usepackage{algpseudocode}
\usepackage{arydshln}
\usepackage[accsupp]{axessibility}

\newcommand{\beginsupp}{
        \setcounter{table}{0}
        \renewcommand{\thetable}{S\arabic{table}}
        \setcounter{figure}{0}
        \renewcommand{\thefigure}{S\arabic{figure}}
        \setcounter{section}{0}
        \renewcommand{\thesection}{\Alph{section}}
     }

\usepackage[pagebackref,breaklinks,colorlinks]{hyperref}

\usepackage[capitalize]{cleveref}
\crefname{section}{Sec.}{Secs.}
\Crefname{section}{Section}{Sections}
\Crefname{table}{Table}{Tables}
\crefname{table}{Tab.}{Tabs.}

\def\cvprPaperID{9104} 
\def\confName{CVPR}
\def\confYear{2023}

\begin{document}

\title{Extracting Class Activation Maps from Non-Discriminative Features as well}

\author{Zhaozheng Chen\\
Singapore Management University\\
{\tt\small zzchen.2019@phdcs.smu.edu.sg}
\and
Qianru Sun\\
Singapore Management University\\
{\tt\small qianrusun@smu.edu.sg}
}

\maketitle

\maketitle

\begin{abstract}
Extracting class activation maps (CAM) from a classification model often results in poor coverage on foreground objects, i.e., only the discriminative region (e.g., the ``head'' of ``sheep'') is recognized and the rest (e.g., the ``leg'' of ``sheep'') mistakenly as background. The crux behind is that the weight of the classifier (used to compute CAM) captures only the discriminative features of objects. We tackle this by introducing a new computation method for CAM that explicitly captures non-discriminative features as well, thereby expanding CAM to cover whole objects. Specifically, we omit the last pooling layer of the classification model, and perform clustering on all local features of an object class, where ``local'' means ``at a spatial pixel position''. We call the resultant $K$ cluster centers \emph{local prototypes} --- represent local semantics like the ``head'', ``leg'', and ``body'' of ``sheep''.  Given a new image of the class, we compare its unpooled features to every prototype, derive $K$ similarity matrices, and then aggregate them into a heatmap (i.e., our CAM). Our CAM thus captures all local features of the class without discrimination. We evaluate it in the challenging tasks of weakly-supervised semantic segmentation (WSSS), and plug it in multiple state-of-the-art WSSS methods, such as MCTformer~\cite{mctformer} and AMN~\cite{amn}, by simply replacing their original CAM with ours. Our extensive experiments on standard WSSS benchmarks (PASCAL VOC and MS COCO) show the superiority of our method: consistent improvements with little computational overhead. Our code is provided at this \href{https://github.com/zhaozhengChen/LPCAM}{link}.
\end{abstract}
\section{Introduction}
\label{sec_intro}
\begin{figure*}[ht]
    \centering
    \includegraphics[width=0.95\linewidth]{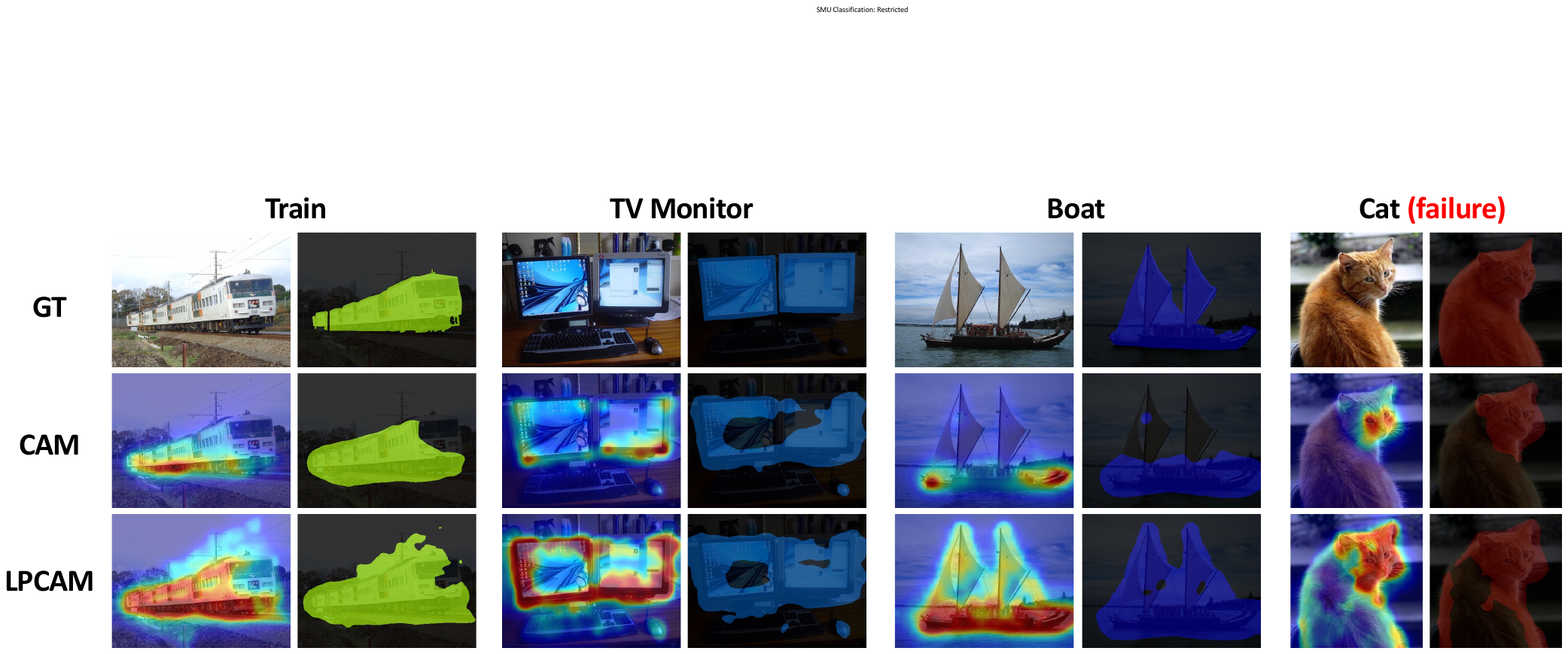}
    \vspace{-2mm}
    \caption{The CAM of image $\bm{x}$ is computed by $f(\bm{x})\cdot\mathbf{w}_c$, where $f(\bm{x})$ is the feature map block (before the last pooling layer of the multi-label classification model) and $\mathbf{w}_c$ denotes the classifier weights of class $c$. (a) Input images. (b) CAMs generated from the classifier weights of ground truth class. (c) CAMs generated from the classifier weights of confusing classes. (d) LPCAM (our method).}
    \label{fig_cam_examples}
\end{figure*}

Extracting CAM~\cite{cam} from classification models is the essential step for training semantic segmentation models when only image-level labels are available, i.e., in the WSSS tasks~\cite{conta,advcam,rib,recam,amn}. More specifically, the general pipeline of WSSS consists of three steps:
1)~training a multi-label classification model with the image-level labels; 2)~extracting CAM of each class to generate a 0-1 mask (usually called seed mask), often with a further step of refinement to generate pseudo mask~\cite{sec,irn}; and 3)~taking all-class pseudo masks as pseudo labels to train a semantic segmentation model in a fully-supervised fashion~\cite{deeplabv2,deeplabv3+,upernet}. It is clear that the CAM in the first step determines the performance of the final semantic segmentation model. However, the conventional CAM and its variants often suffer from the poor coverage of foreground objects in the image, i.e., a large amount of object pixels are mistakenly recognized as background, as demonstrated in Figure~\ref{fig_cam_examples}(a) where only few pixels are activated in warm colors.

We point out that this locality is due to the fact that CAM is extracted from a discriminative model. The training of such model naturally discards the non-discriminative regions which confuse the model between similar as well as highly co-occurring object classes. This is a general problem of discriminative models, and is particularly obvious when the number of training classes is small~\cite{conta,adv_erasing,adv_erasing2}. To visualize the evidence, we use the classifier weights of confusing classes, e.g., ``car'', ``train'' and ``person'', to compute CAMs for the ``bus'' image, and show them in Figure~\ref{fig_cam_examples}(b). We find from (a) and (b) that the heating regions in ground truth class and confusing classes are complementary. E.g., the upper and frontal regions (on the ``bus'' image) respectively heated in the CAMs of ``car'' and ``train'' (confusing classes) are missing in the CAM of ``bus'' (ground truth), which means the classifier de-activates those regions for ``bus'' as it is likely to recognize them as ``car'' or ``train''.

Technically, for each class, we use two factors to compute CAM: 1) the feature map block after the last conv layer, and 2) the weight of the classifier for that class.  As aforementioned, the second factor is often biased to discriminative features. Our intuition is to replace it with a non-biased one. The question becomes how to derive a \emph{non-biased} classifier from a biased classification model, where \emph{non-biased} means representing all local semantics of the class. We find the biased classifier is due to incomplete features, i.e., only discriminative local features fed into the classifier, and the non-discriminative ones are kicked out by the global average pooling (GAP) after the last conv layer. To let the model pay attention to non-discriminative features as well, we propose to omit GAP, derive a prototype-based classifier by clustering all local features (collected across all spatial locations on the feature map blocks of all training samples in the class) into $K$ local prototypes each representing a local semantic of the class. In Section~\ref{sec_method}, we give a detailed justification that this prototype-based classifier is able to capture both discriminative and non-discriminative features.

Then, the question is how to use local prototypes on the feature map block (i.e., the 1st factor) to generate CAM. We propose to apply them one-by-one on the feature map block to generate $K$ similarity maps (e.g., by using cosine distance), each aiming to capture the local regions that contain similar semantics to one of the prototypes. We highlight that this ``one-by-one'' is important to preserve non-discriminative regions as the normalization on each similarity map is independent. We provide a detailed justification from the perspective of normalization in Section~\ref{sec_justfication}. Then, we average across all normalized similarity maps to get a single map---we call Local Prototype CAM (LPCAM). In addition, we extend LPCAM by using the local prototypes of contexts as well. We subtract the context similarity maps (computed between the feature map block and the clustered context prototypes) from LPCAM. The idea is to remove the false positive pixels (e.g., the ``rail'' of ``train''~\cite{ood}). Therefore, our LPCAM not only captures the missing local features of the object but also mitigates the spurious features caused by confusing contexts.

LPCAM is a new operation to compute class activation maps based on clustered local prototypes. In principle, it can be taken as a generic substitute of the conventional CAM in CAM-based WSSS methods. To evaluate LPCAM on different WSSS methods (as they use different backbones, pre-training strategies or extra data), we conduct extensive experiments by plugging it in multiple methods: the popular refinement method IRN~\cite{irn}, the top-performing AMN~\cite{amn}, the saliency-map-based EDAM~\cite{edam}, and the transformer-arch-based MCTformer~\cite{mctformer},
on two popular benchmarks of semantic segmentation, PASCAL VOC 2012~\cite{voc} and MS COCO 2014~\cite{mscoco}.

\noindent
\textbf{Our Contributions} in this paper are thus two-fold. 
1)~A novel method LPCAM that leverages non-discriminative local features and context features (in addition to discriminative ones) to generate class activation maps with better coverage on the complete object.
2) Extensive evaluations of LPCAM by plugging it in multiple WSSS methods, on two popular WSSS benchmarks.
\section{Related Works}
\label{sec_related}

Image classification models are optimized to capture only the discriminative local regions (features) of objects, leading to the poor coverage of its CAM on the objects. While the ideal pseudo labels (extended from CAM) to train WSSS models should cover all parts of the object regardless of dicriminativeness. To this end, many efforts have been made in the field of WSSS. Below, we introduce only the variants for seed generation and mask refinement.

\noindent
\textbf{Seed Generation}
One direction is the erasing-based methods. AE-PSL~\cite{adv_erasing} is an adversarial erasing strategy running in an iterative manner. It masks out the discriminative regions in the current iteration, to explicitly force the model to discover new regions in the next iteration. ACoL~\cite{adv_erasing2} is an improved method. It is based on an end-to-end learning framework composed of two branches. One branch applies a feature-level masking on the other. However, these two methods have an over-erasing problem, especially for small objects. CSE~\cite{cse} is a class-specific erasing method. It masks out a random object class based on CAM and then explicitly penalizes the prediction of the erased class. In this way, it gradually approaches to the boundary of the object on the image. It can not only discover more non-discriminative regions but also alleviate the over-erasing problem (of the above two methods) since it also penalizes over-erased regions. However, erasing-based methods have the low-efficiency problem as they have to feed forward an image multiple times.

Besides of erasing, there are other advanced methods recently.
RIB~\cite{rib} interpreted the poor coverage problem of CAM in an information bottleneck principle. It re-trains the multi-label classifier by omitting the activation function in the last layer to encourage the information transmission of information non-discriminative regions to the classification. Jiang et al.~\cite{l2g} empirically observed that classification models can discover more discriminative regions when taking local image patches rather than whole input image as input. They proposed a local-to-global attention transfer method contains a local network that produces local attentions with rich object details for local views as well as a global network that receives the global view as input and aims to distill the discriminative attention knowledge from the local network. Some other researchers explore to utilizing contrastive learning~\cite{rca,ppc}, graph neural network~\cite{group}, and self-supervised learning~\cite{sub_category,seam} to discover more non-discriminative regions.

Compared to aforementioned methods, our method have the following advantages:
1) it does not require additional training on the basis of CAM;
and 2) it can be taken as a generic substitute of the conventional CAM and plugged into many CAM-based WSSS frameworks.

\noindent
\textbf{Mask Refinement}
One category of refinement methods~\cite{psa,irn,auxiliary,bes} propagate the object regions in the seed to semantically similar pixels in the neighborhood. It is achieved by the random walk~\cite{randomwalk} on a transition matrix where each element is an affinity score. The related methods have different designs of this matrix. PSA~\cite{psa} is an AffinityNet to predict semantic affinities between adjacent pixels. IRN~\cite{irn} is an inter-pixel relation network to estimate class boundary maps based on which it computes affinities. Another method is BES~\cite{bes} that learns to predict boundary maps by using CAM as pseudo ground truth. All these methods introduced additional network modules to vanilla CAM. Another category of refinement methods~\cite{dsrg,ooa,ficklenet,edam,eps} utilize saliency maps~\cite{saliency1,saliency2} to extract background cues and combine them with object cues. EPS~\cite{eps} proposed a joint training strategy to combine CAM and saliency maps. EDAM~\cite{edam} introduced a post-processing method to integrate the confident areas in the saliency map into CAM. Our LPCAM is orthogonal to them and can be plugged into those methods.
\section{Method}
\label{sec_method}
\begin{figure*}[ht]
    \centering
    \includegraphics[width=0.99\linewidth]{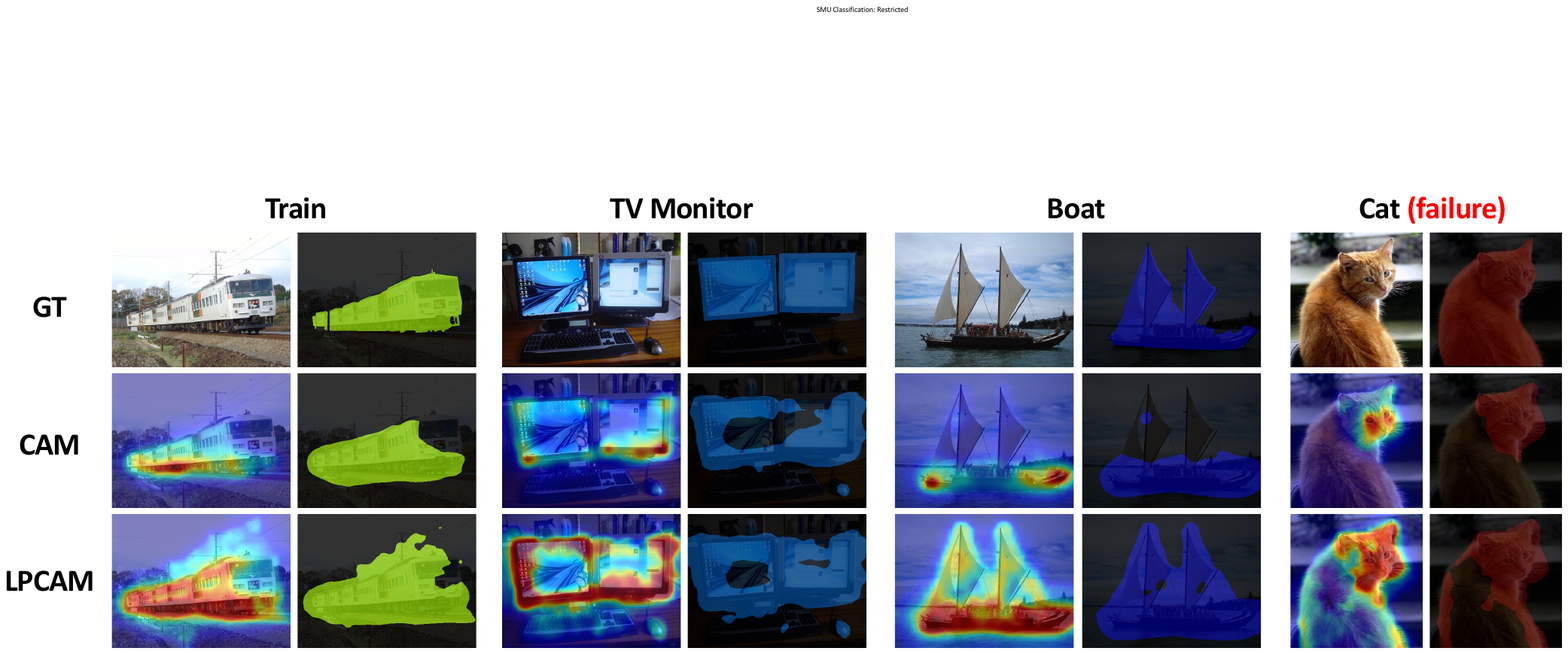}
    \vspace{-2mm}
    \caption{The pipeline for generating LPCAM. (a) Generating local prototypes from all training images of an individual class, e.g., ``sheep''. 
    (b) Generating LPCAM by \emph{sliding} both class prototypes and context prototypes over the feature block of the input image, and \emph{aggregating} the obtained similarity maps.
    In the gray block, the generation process of CAM is shown for comparison. }
    \vspace{-2mm}
    \label{fig_framework}
\end{figure*}
In Section~\ref{sec_pipeline}, we introduce the pipeline of generating LPCAM using a collection of class-wise prototypes including class prototypes and context prototypes, without any re-training on the classification model. The step-by-step illustration is shown in Figure~\ref{fig_framework}, demonstrating the steps of generating local prototypes from all images of a class and using these prototypes to extract LPCAM for each single image. In Section~\ref{sec_justfication}, we justify 1) the advantages of using clustered local prototypes in LPCAM; and 2) the effectiveness of LPCAM from the perspective of map normalization. 

\subsection{LPCAM Pipeline}
\label{sec_pipeline}

\noindent
\textbf{Backbone and Features.} 
We use a standard ResNet-50~\cite{resnet} as the network backbone (i.e., feature encoder) of the multi-label classification model to extract features, following related works~\cite{irn,conta,advcam,rib,recam,amn}. Given an input image $\bm{x}$, and its multi-hot class label $\bm{y}\in\{0,1\}^{N}$,  we denote the output of the trained feature encoder as $f(\bm{x}) \in \mathbb{R}^{W \times H \times C}$. $C$ denotes the number of channels, $H$ and $W$ are the height and width, respectively, and $N$ is the total number of foreground classes in the dataset.

\noindent
\textbf{Extracting CAM.}
Before clustering local prototypes of class as well as context, we need the rough location information of foreground and background. We use the conventional CAM to achieve this. We extract it for each individual class $n$ given the feature $f(\bm{x})$ and the corresponding classifier weights $\mathbf{w}_n$ in the FC layer, as follows,
\begin{equation} \label{eq:cam}
    \operatorname{CAM}_n(\bm{x})=
    \frac{\operatorname{ReLU}\left(\bm{A}_n\right)}{\max \left(\operatorname{ReLU}\left(\bm{A}_n\right)\right)},
    \quad
    \bm{A}_n=\mathbf{w}_{n}^{\top}f(\bm{x}).
\end{equation}

\noindent
\textbf{Clustering.}
We perform clustering for every individual class. Here we discuss the details for class $n$. Given an image sample $\bm{x}$ of class $n$, we divide the feature block $f(\bm{x})$ spatially into two sets, $\mathcal{F}$ and $\mathcal{B}$, based on CAM:
\begin{equation}\label{eq:cam_fg_bg}
    f(\bm{x})^{i,j} \in 
        \begin{cases}
            \mathcal{F}, & \text { if } \operatorname{CAM}_n^{i,j}(\bm{x}) \geq \tau \\ 
            \mathcal{B}, & \text { otherwise }
        \end{cases}
\end{equation}
where $f(\bm{x})^{i,j} \in \mathbb{R}^{C}$ denotes the local feature at spatial location $(i,j)$. $\tau$ is the threshold to generate a 0-1 mask from $\operatorname{CAM}_n(\bm{x})$. $\mathcal{F}$ denotes the set of foreground local features, and $\mathcal{B}$ for the set of background (context) local features.

Similarly, we can collect $\mathcal{F}$ to contain the foreground features of all samples\footnote{We use a random subset of samples for each class in the real implementation, to reduce the computation costs of clustering.} in class $n$, and $\mathcal{B}$ for all background features, where the subscript $n$ is omitted for brevity. After that, we perform K-Means clustering, respectively, for $\mathcal{F}$ and $\mathcal{B}$, to obtain $K$ class centers in each of them, where $K$ is a hyperparameter. We denote the foreground cluster centers as $\mathbf{F}=\{\mathbf{F}_{1}, \cdots, \mathbf{F}_{K}\}$ and the background cluster centers as $\mathbf{B}=\{\mathbf{B}_{1}, \cdots, \mathbf{B}_{K}\}$.

\noindent
\textbf{Selecting Prototypes.}
The masks of conventional CAM are not accurate or complete, e.g., background features could be grouped into $\mathcal{F}$. To solve this issue, we need an ``evaluator'' to check the eligibility of cluster centers to be used as prototypes. The intuitive way is to use the classifier $\mathbf{w}_n$ as an auto ``evaluator'': using it to compute the prediction score of each cluster center $\mathbf{F}_i$ in $\mathbf{F}$ by:
\begin{equation}
    \bm{z}_i = \frac{\exp(\mathbf{F}_i \cdot \mathbf{w}_n)}{\sum_{j} \exp (\mathbf{F}_i \cdot \mathbf{w}_j)}.
\end{equation}
Then, we select those centers with high confidence: $\bm{z}_i>\mu_f$, where $\mu_f$ is a threshold---usually a very high value like $0.9$. 
We denote selected ones as $\mathbf{F'}=\{\mathbf{F'}_{1}, \cdots, \mathbf{F'}_{K'_1}\}$. Intuitively, confident predictions indicate strong local features, i.e., prototypes, of the class.

Before using these local prototypes to generate LPCAM, we highlight that in our implementation of LPCAM, we not only \emph{preserve} the non-discriminative features but also \emph{suppress} strong context features (i.e., false positive),
as the extraction and application of context prototypes are convenient---similar to class prototypes but in a reversed manner. We elaborate these in the following.
For each $\mathbf{B}_i$ in the context cluster center set $\mathbf{B}$, we apply the same method (as for $\mathbf{F}_i$) to compute a prediction score:
\begin{equation}
    \bm{z}_i = \frac{\exp(\mathbf{B}_i \cdot \mathbf{w}_n)}{\sum_{j} \exp (\mathbf{B}_i \cdot \mathbf{w}_j)}.
\end{equation}
Intuitively, if the model is well-trained on class labels, its prediction on context features should be extremely low.
Therefore, we select the centers with $\bm{z}_i<\mu_b$ (where $\mu_b$ is usually a value like $0.5$), and denote them as $\mathbf{B'}=\{\mathbf{B'}_{1}, \cdots, \mathbf{B'}_{K'_2}\}$. It is worth noting that our method is not sensitive to the values of the hyperparameters $\mu_f$ and $\mu_b$, given reasonable ranges, e.g., $\mu_f$ should have a large value around $0.9$. We show an empirical validation for this in Section~\ref{sec_exper}.

\noindent
\textbf{Generating LPCAM.}
Each of the prototypes represents a local visual pattern in the class: $\mathbf{F'}_{i}$ for class-related pattern (e.g., the ``leg'' of ``sheep'' class) and $\mathbf{B'}_{i}$ for context-related pattern (e.g., the ``grassland'' in ``sheep'' images) where the context often correlates with the class. Here we introduce how to apply these prototypes on the feature map block to generate LPCAM. LPCAM can be taken as a substitute of CAM. In Subsection~\ref{sec_justfication}, we will justify why LPCAM is superior to CAM from two perspectives: Global Average Pooling (GAP) and Normalization.

For each prototype, we slide it over all spatial positions on the feature map block, and compute its similarity to the local feature at each position. We adopt cosine similarity as we used it for K-Means. In the end, we get a cosine similarity map between prototype and feature. After computing all similarity maps (by sliding all local prototypes), we aggregate them as follows,
\begin{equation}
    \begin{aligned}
         &\bm{FG}_n=\frac{1}{K'_1}\sum_{\mathbf{F'_i}\in\mathbf{F'}}sim(f(\bm{x}),\mathbf{F'_i}), \\
        & \bm{BG}_n=\frac{1}{K'_2}\sum_{\mathbf{B'_i}\in\mathbf{B'}}sim(f(\bm{x}),\mathbf{B'_i}),
    \label{eq:fg_bg}
    \end{aligned}
\end{equation}
where $sim()$ denotes cosine similarity. As $sim()$ value is always within the range of $[-1, 1]$, each pixel on the maps of $\bm{FG}_n$ and $\bm{BG}_n$ has a normalized value, i.e., $\bm{FG}_n$ and $\bm{BG}_n$ are normalized. Intuitively, $\bm{FG}_n$ highlights the class regions in the input image correlated to the $n$-th prototype, while $\bm{BG}_n$ highlights the context regions. The former needs to be preserved and the latter (e.g., pixels highly correlated to backgrounds) should to be removed. Therefore, we can formulate LPCAM as follows:
\begin{equation}
    \begin{aligned}
        &\operatorname{LPCAM}_n(\bm{x})=
        \frac{\operatorname{ReLU}\left(\bm{A}_n\right)}{\max \left(\operatorname{ReLU}\left(\bm{A}_n\right)\right)}, \\
        &\bm{A}_n= \bm{FG}_n - \bm{BG}_n,
    \label{eq:LPCAM}
    \end{aligned}
\end{equation}
where the first formula is the application of the maximum-value based normalization (the same as in CAM).

\subsection{Justifications}
\label{sec_justfication}
\begin{figure}[ht]
    \centering
    \includegraphics[width=0.99\linewidth]{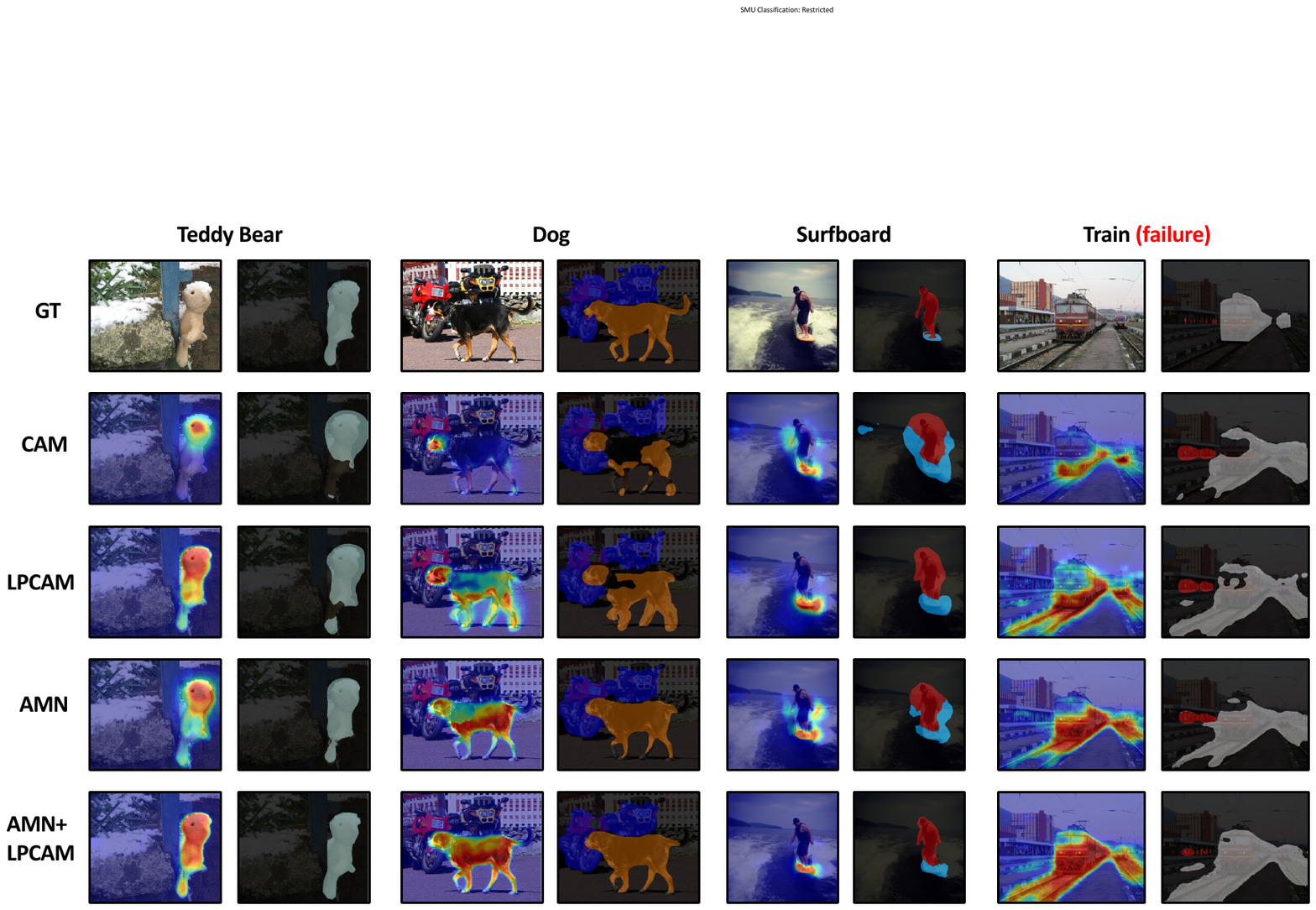}
    \vspace{-2mm}
    \caption{Justifying the advantages of LPCAM over CAM from two perspectives: (a) clustering and (b) normalization. For simplicity, in (a), we consider only three local regions on a ``bird'' image: $\bm{x_1}$: ``head'', $\bm{x_2}$: ``tail'', $\bm{x_3}$: ``sky''. In (b), we assume two class prototypes (``head'' and ``tail''), and one context prototype (``sky'') are selected after local feature clustering.}
    \vspace{-2mm}
    \label{fig_justification}
\end{figure}
We justify the effectiveness of LPCAM from two perspectives: clustering and normalization. We use the ``bird'' example shown in Figure~\ref{fig_justification}. In (a), we consider only three local regions ($\bm{x_1}$, $\bm{x_2}$ and $\bm{x_3}$), for simplicity. Their semantics are respectively: $\bm{x_1}$ as ``head'' (a discriminative object region), $\bm{x_2}$ as ``tail'' (a non-discriminative region), and $\bm{x_3}$ as ``sky'' (a context region). We suppose $f(\bm{x_1})$, $f(\bm{x_2})$, and $f(\bm{x_3})$ are 3-dimensional local features (2048-dimensional features in our real implementation) respectively extracted from the three regions, where the three dimensions represent the attributes of ``head'', ``tail'' and ``sky'', respectively. The discriminativeness is reflected as follows. First, $f(\bm{x_1})$ extracted on the head region $\bm{x_1}$ has a significantly higher value of the first dimension than $f(\bm{x_2})$ and $f(\bm{x_3})$. Second, $f(\bm{x_2})$ has the highest value on the second dimension but this value is lower than the first dimension of $f(\bm{x_1})$, because ``tail'' ($\bm{x_2}$) is less discriminative than ``head'' ($\bm{x_1}$) for recognizing ``bird''. In (b), we assume three local class prototypes (``head'', ``tail'', and ``sky'') are selected.

\noindent
\textbf{Clustering.} 
As shown in Figure~\ref{fig_justification}(a), 
\textbf{in LPCAM}, $\bm{x_1}$ and $\bm{x_2}$ go to different clusters. This is determined by their dominant feature dimensions, i.e., the first dimension in $f(\bm{x_1})$ and the second dimension in $f(\bm{x_2})$. Given all samples of ``bird'', their features clustered into the ``head'' cluster all have high values in the first dimension, and features in the ``tail'' have high values in the second dimension. The centers of these clusters are taken as local prototypes and equally used for generating LPCAM. Sliding each of prototypes over the feature map block (of an input image) can highlight the corresponding local region. The intuition is each prototype works like a spatial-wise filter that amplifies similar regions and suppresses dissimilar regions.

However, \textbf{in CAM}, the heatmap computation uses the classifier weights biased on discriminative dimensions\footnote{We empirically validate this in the supplementary materials.}. It is because the classifier is learned from the global average pooling features, e.g., $f_{GAP}=\frac{1}{3}(f(\bm{x_1})+f(\bm{x_2})+f(\bm{x_3}))=[12,5,4]$ biased to the ``head'' dimension. As a result, only discriminative regions (like ``head'' for the class of ``bird'') are highlighted on the heatmap of CAM.

\noindent
\textbf{Normalization.} 
We justify the effectiveness of \textbf{LPCAM} by presenting the normalization details in Figure~\ref{fig_justification}(b). We denote the two class prototypes (``head'' and ``tail'') as $\mathbf{F'_1}, \mathbf{F'_2}$ and the context prototype (``sky'') as $\mathbf{B'_1}$. Based on Eq.~\ref{eq:fg_bg} and Eq.~\ref{eq:LPCAM}, we have $\bm{A}(\bm{x})=\frac{1}{2}\sum_{i=1}^{2}sim(f(\bm{x}),\mathbf{F'_i}) - sim(f(\bm{x}),\mathbf{B'_1})$, where $\bm{x}$ denotes any local region. For simplicity, we use the first term for explanation: the prototype ``head'' $\mathbf{F'_1}$ has the highest similarity ($1.0$) to region $\bm{x_1}$ and the prototype ``tail'' $\mathbf{F'_2}$ has the highest similarity ($0.9$) to $\bm{x_2}$. $0.9$ and $1.0$ are very close. After the final maximum-value based normalization (as in the first formula of Eq.~\ref{eq:LPCAM}), they become $1.00$ and $0.83$, i.e., only a small gap between discriminative ($\bm{x_1}$) and non-discriminative ($\bm{x_2}$) regions. However, \textbf{CAM} (Eq.~\ref{eq:cam}) uses $\mathbf{w}^{\top}f(\bm{x})$, resulting a much higher activation value of $\bm{x_1}$ than $\bm{x_2}$, as $\mathbf{w}$ is obviously biased to ``head'', i.e., $130$ vs. $35$ (``tail''). After the final maximum-value based normalization, there is no change to this bias: $1.00$ and $0.27$---a large gap. In other words, the non-discriminative feature is closer to background $\bm{x_3}$, making the boundary between foreground and background blurry, and hard to find a threshold to separate them.

One may argue that ``separate $\bm{x_2}$ and $\bm{x_3}$'' can be achieved in either CAM or LPCAM if the threshold is carefully selected in each method. However, it is not realistic to do such ``careful selection'' for every input image. The general way in WSSS is to use a common threshold for all images. Our LPCAM makes it easier to find such a threshold, since its heatmap has a much clearer boundary between foreground and background. We conduct a threshold sensitivity analysis in experiments to validate this.

\section{Experiments}
\label{sec_exper}

\subsection{Datasets and Implementation Details}
\label{sec_datasets}

\noindent
\textbf{Datasets} are the commonly used WSSS datasets: PASCAL VOC 2012~\cite{voc} and MS~COCO 2014~\cite{mscoco}. PASCAL VOC 2012 contains $20$ foreground object categories and $1$ background category with $1,464$ \texttt{train} images, $1,449$ \texttt{val} images, and $1,456$ \texttt{test} images. Following related works~\cite{irn,recam,advcam,amn,ppc,rca}, we use the enlarged training set with $10,582$ training images provided by SBD~\cite{voc_aug}. MS~COCO 2014 dataset consists of $80$ foreground categories and $1$ background category, with $82,783$ and $40,504$ images in \texttt{train} and \texttt{val} sets, respectively. 

\noindent
\textbf{Evaluation Metrics.}
To evaluate the quality of seed mask and pseudo mask, we first generate them for every image in the \texttt{train} set and then use the ground truth masks to compute mIoU. For semantic segmentation, we train the segmentation model, use it to predict masks for the images in \texttt{val} and \texttt{test} sets, and compute mIoU based on ground truth masks.

\noindent
\textbf{Implementation Details.}
We follow \cite{irn,conta,recam,advcam,amn} to use ResNet-50~\cite{resnet} pre-trained on ImageNet~\cite{imagenet} as the backbone of multi-label classification model. For fair comparison with related works, we also follow \cite{seam,ppc,rca} to use WideResNet-38~\cite{wresnet} as backbone and follow EDAM~\cite{edam} to use saliency maps~\cite{saliency} to refine CAM. 

\noindent
\emph{Extra Hyperparameters.}
For K-Means clustering, we set $K$ as $12$ and $20$ for VOC and MS~COCO, respectively. The threshold $\tau$ in Eq.~\ref{eq:cam_fg_bg} is set to $0.1$ for VOC and $0.25$ for MS~COCO. For the selection of prototypes, $\mu_f$ is set to $0.9$ on both datasets, and $\mu_b$ is $0.9$ and $0.5$ on VOC and MS~COCO, respectively. We conduct the sensitivity analysis on these four hyperparameters to show that LPCAM is not sensitive to any of them.

\noindent
\emph{Common Hyperparameter (in CAM methods).}
The hard threshold used to generate 0-1 seed mask is $0.3$ for LPCAM on both datasets. Please note that we follow previous works~\cite{conta,advcam,rib,recam,amn,ppc} to select this threshold by using the ground truth masks in the training set as ``validation''.

\noindent
\emph{Time Costs.}
In K-Means clustering, we use all \texttt{train} set on VOC and sample $100$ images per class on MS~COCO (to control time costs). If taking the time cost of training a multi-label classification model as unit $1$, our extra time cost (for clustering) is about $0.9$ and $1.1$ on VOC and MS~COCO, respectively. 

\noindent
\emph{For Semantic Segmentation.}
When using DeepLabV2~\cite{deeplabv2} for semantic segmentation, we follow the common settings \cite{irn,recam,advcam,rib,amn,ppc} as follows.
The backbone of DeepLabV2 model is ResNet-101~\cite{resnet} and is pre-trained on ImageNet\cite{imagenet}. We crop each training image to the size of $321\times 321$ and use horizontal flipping and random crop for data augmentation. We train the model for $20k$ and $100k$ iterations on VOC and MS COCO, respectively, with the respective batch size of $5$ and $10$. The weight decay is set to $5$e-$4$ on both datasets and the learning rate is $2.5$e-$4$ and $2$e-$4$ on VOC and MS~COCO, respectively. 

When using UperNet, we follow ReCAM~\cite{recam}. We resize the input images to $2,048\times 512$ with a ratio range from $0.5$ to $2.0$, and then crop them to $512\times 512$ randomly. Data augmentation includes horizontal flipping and color jitter. We train the models for $40k$ and $80k$ iterations on VOC and MS~COCO datasets, respectively, with a batch size of 16. We deploy AdamW~\cite{adamw} solver with an initial learning rate $6e^{-5}$ and weight decay as $0.01$. The learning rate is decayed by a power of $1.0$ according to polynomial decay schedule.

\subsection{Results and Analyses}

\noindent
\textbf{Ablation Study.}
We conduct an ablation study on the VOC dataset to evaluate the two terms of LPCAM in Eq.~\ref{eq:LPCAM}: foreground term $\bm{FG}_n$ and background term $\bm{BG}_n$ that accord to class and context prototypes, respectively. In Table~\ref{table_ablation}, we show the mIoU results (of seed masks), false positive (FP), false negative (FN), precision, and recall. We can see that our methods of using class prototypes (LPCAM-F and LPCAM) greatly improve the recalls---$11.4\%$ and $12.0\%$ higher than CAM, reducing the rates of FN a lot. This validates the ability of our methods to capture non-discriminative regions of the image. We also notice that LPCAM-F increases the rate of FP over CAM. The reason is that confusing context features (e.g., ``railroad'' for ``train'') may be wrongly taken as class features. Fortunately, when we explicitly resolve this issue by applying the negative context term $-\bm{BG}_n$ in LPCAM, this rate can be reduced (by $3.3\%$ for VOC), and the overall performance (mIoU) can be improved (by $2.8\%$ for VOC). We are thus confident to take LPCAM as a generic substitute of CAM in WSSS methods (see empirical validations below). 

\setlength{\tabcolsep}{2.2mm}{
\renewcommand\arraystretch{1.1}
\begin{table}[ht]
    \centering
    \scalebox{0.8}{
      \begin{tabular}{lccccc}
        \toprule
                      & FP    & FN    & mIoU  & Prec. & Recall\\ 
        \midrule
        \texttt{CAM}           & 26.5    & 26.2  & 48.8  & 65.0  & 65.2  \\
        \texttt{LPCAM-F}      & 33.1\scriptsize{+6.6} & 16.2\scriptsize{-10.0} & 52.1\scriptsize{+3.3}  & 61.3\scriptsize{-3.7}  & 76.6\scriptsize{+11.4}  \\
        \texttt{LPCAM}         & 29.8\scriptsize{+3.3} & 16.7\scriptsize{-9.5} & 54.9\scriptsize{+6.1}  & 64.9\scriptsize{-0.1}  & 77.2\scriptsize{+12.0}  \\
        \bottomrule
      \end{tabular}
    }
    \vspace{-2mm}
    \caption{An ablation study on VOC dataset. ``-F'' denotes only the ``Foreground'' term $\bm{FG}_n$ is used in Eq.~\ref{eq:LPCAM}. \emph{Please refer to the supplementary materials for the results on MS~COCO.}}
    \label{table_ablation}
\end{table}
}
\setlength{\tabcolsep}{1.9mm}{
\renewcommand\arraystretch{1}
\begin{table}[ht]
  \centering
  \scalebox{0.95}{
  \begin{tabular}{llcccc}
    \toprule
    &\multirow{2}*{Methods}& \multicolumn{2}{c}{Seed Mask} & \multicolumn{2}{c}{Pseudo Mask}  \\
    \cmidrule(r){3-4}\cmidrule(r){5-6}
    && \texttt{CAM} & \texttt{LPCAM} & \texttt{CAM} & \texttt{LPCAM}  \\
    \midrule
    \multirow{4}*{\rotatebox{90}{\small{VOC}}} &IRN~\cite{irn}  & 48.8  & 54.9\scriptsize{+6.1}   & 66.5  & 71.2\scriptsize{+4.7}    \\
    ~&EDAM~\cite{edam}                   & 52.8  & 54.9\scriptsize{+2.1}   & 68.1 & 69.6\scriptsize{+1.5}    \\
    ~&MCTformer~\cite{mctformer}        & 61.7  & 63.5\scriptsize{+1.8} & 69.1  & 70.8\scriptsize{+1.7} \\
    ~&AMN~\cite{amn}                 & 62.1  & 65.3\scriptsize{+3.2} &72.2& 72.7\scriptsize{+0.5}   \\
    \midrule
    \multirow{2}*{\rotatebox{90}{\small{COCO}}} & IRN~\cite{irn} & 33.1  & 35.8\scriptsize{+2.7}     & 42.5  & 46.8\scriptsize{+4.3}    \\
    ~&AMN~\cite{amn}                        & 40.3  & 42.5\scriptsize{+2.2}    & 46.7  & 47.7\scriptsize{+1.0}    \\
    \bottomrule
  \end{tabular}}
  \vspace{-0.2cm}
  \caption{Taking LPCAM as a substitute of CAM in state-of-the-art WSSS methods. Except MCTformer~\cite{mctformer} using DeiT-S~\cite{deit}, other methods all use ResNet-50 as feature extractor.}
  \label{table_plugin}
  \vspace{-0.4cm}
\end{table}
}

\begin{figure*}[ht]
    \centering
    \includegraphics[width=0.99\linewidth]{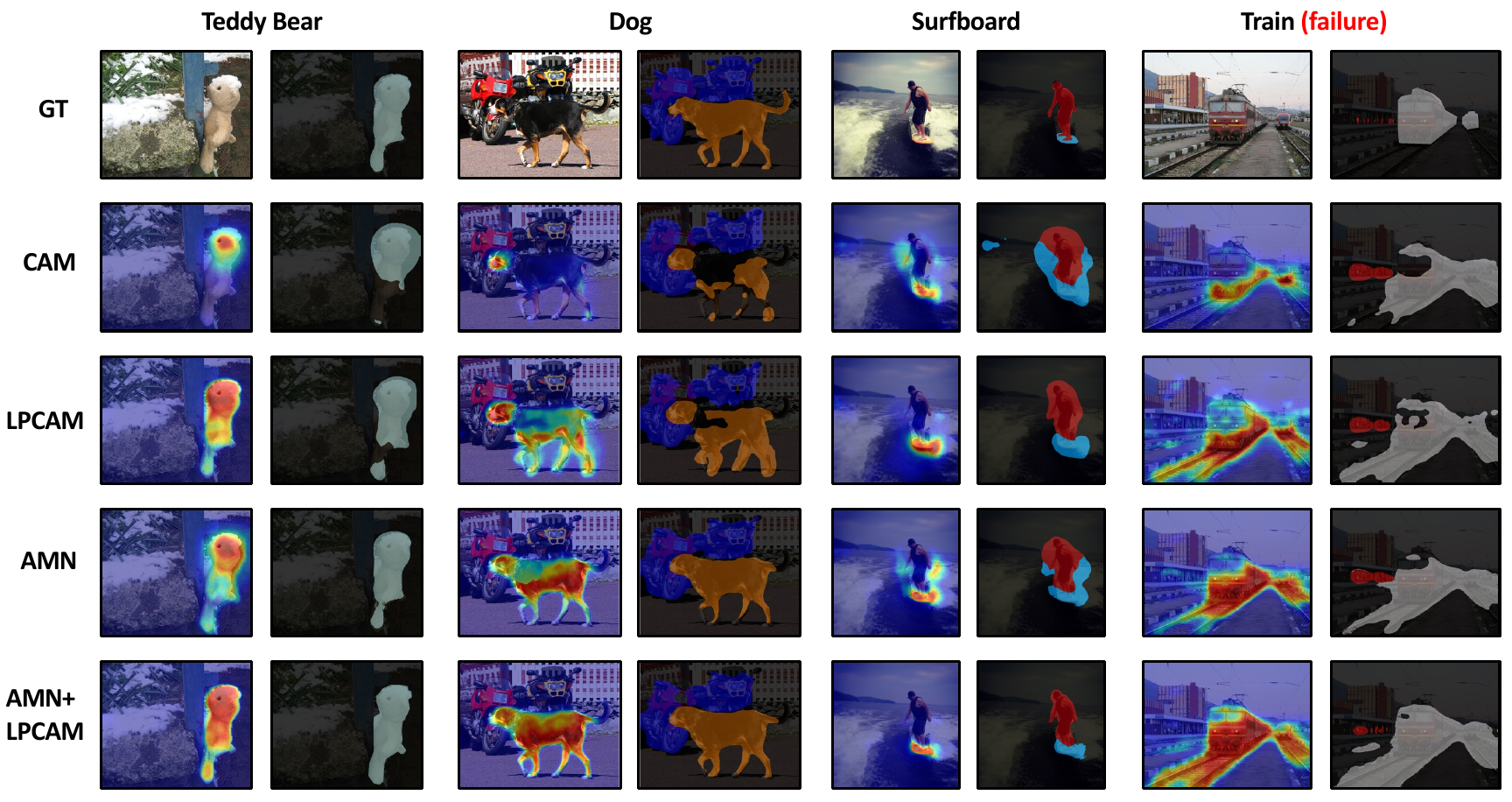}
    \caption{Qualitative results on MS~COCO. In each example pair, the left is heatmap and the right is seed mask. \emph{Please refer to the supplementary materials for the qualitative results on VOC.}}
    \vspace{-2mm}
    \label{fig_vis}
\end{figure*}
\setlength{\tabcolsep}{1.2mm}{
\renewcommand\arraystretch{1}
\begin{table*}
  \centering
  \begin{tabular}{lcccccccccccc}
    \toprule
    \multirow{4}*{Methods}& \multicolumn{8}{c}{VOC}& \multicolumn{4}{c}{MS~COCO} \\
    \cmidrule(r){2-9}\cmidrule(r){10-13}
    & \multicolumn{4}{c}{DeepLabV2} &\multicolumn{4}{c}{UperNet-Swin}& \multicolumn{2}{c}{DeepLabV2} &\multicolumn{2}{c}{UperNet-Swin}\\
    \cmidrule(r){2-5}\cmidrule(r){6-9}\cmidrule(r){10-11}\cmidrule(r){12-13}
    & \multicolumn{2}{c}{\texttt{CAM}} & \multicolumn{2}{c}{\texttt{LPCAM}}& \multicolumn{2}{c}{\texttt{CAM}} & \multicolumn{2}{c}{\texttt{LPCAM}}& \texttt{CAM} &\texttt{LPCAM} & \texttt{CAM} &\texttt{LPCAM} \\
    \cmidrule(r){2-3}\cmidrule(r){4-5}\cmidrule(r){6-7}\cmidrule(r){8-9}\cmidrule(r){10-10}\cmidrule(r){11-11}\cmidrule(r){12-12}\cmidrule(r){13-13}
    & \texttt{val} & \texttt{test} & \texttt{val} & \texttt{test}& \texttt{val} & \texttt{test} & \texttt{val} & \texttt{test} & \texttt{val} & \texttt{val} & \texttt{val} & \texttt{val} \\
    \hline
    IRN~\cite{irn}  & 63.5  & 64.8  & 68.6\scriptsize{+5.1} & 68.7\scriptsize{+3.9} & 65.9  & 67.7  & 71.1\scriptsize{+5.2} &  71.8\scriptsize{+4.1}   & 42.0  & 44.5\scriptsize{+2.5} & 44.0  & 47.0\scriptsize{{+3.0}}  \\
    AMN~\cite{amn}  & 69.5  & 69.6  & 70.1\scriptsize{+0.6} & 70.4\scriptsize{+0.8}    & 71.7    & 71.8    & 73.1\scriptsize{+1.4}    &  73.4\scriptsize{+1.6}   & 44.7  & 45.5\scriptsize{+0.8}    & 47.1    & 48.3\scriptsize{+1.2}    \\
    EDAM\cite{edam} & 70.9$^*$  & 70.6$^*$  & 71.8$^*$\scriptsize{+0.9}  & 72.1$^*$\scriptsize{+1.5}   & 71.2  & 71.0  & 72.7\scriptsize{+1.6}   &  72.5\scriptsize{+1.5}   & 40.6  & 42.1\scriptsize{+1.5} & 41.7  & 43.0\scriptsize{+1.3}    \\
    MCTformer\cite{mctformer} & 71.9$^\dag$  & 71.6$^\dag$  & 72.6$^\dag$\scriptsize{+0.7}  & 72.4$^\dag$\scriptsize{+0.8}   & 70.6  & 70.3  & 72.0\scriptsize{+1.4}    &  72.5\scriptsize{+2.2}   & -  & - & -  & -    \\
    \bottomrule
  \end{tabular}
  \vspace{-3mm}
  \caption{The mIoU results (\%) of WSSS using different segmentation models on VOC and MS~COCO. Seed masks are generated by either CAM or LPCAM, and mask refinement methods are row titles. %
  * denotes the segmentation model (ResNet-101 based DeepLabV2) is pre-trained on MS~COCO. 
  $^\dag$ denotes the segmentation model (WideResNet-38 based DeepLabV2) is pre-trained on VOC.}
  \label{table_seg}
  \vspace{-0.5cm}
\end{table*}
}

\noindent
\textbf{Generality of LPCAM.}
We validate the generality of LPCAM based on multiple WSSS methods, the popular IRN~\cite{irn}, the top-performing AMN~\cite{amn}, the saliency-map-based EDAM~\cite{edam}, and the transformer-arch-based MCTformer~\cite{mctformer}), by simply replacing CAM with LPCAM. Table~\ref{table_plugin} and Table~\ref{table_seg} show the consistent superiority of LPCAM. For example, on the first row of Table~\ref{table_plugin} (plugging LPCAM in IRN), LPCAM outperforms CAM by $6.1\%$ on seed masks and $4.7\%$ on pseudo masks. These margins are almost maintained when using pseudo masks to train semantic segmentation models in Table~\ref{table_seg}. The improvements on the large-scale dataset MS~COCO are also obvious and consistent, e.g., $2.7\%$ and $2.2\%$ for generating seed masks in IRN and AMN, respectively.

\begin{figure}[ht]
    \centering
    \includegraphics[width=0.99\linewidth]{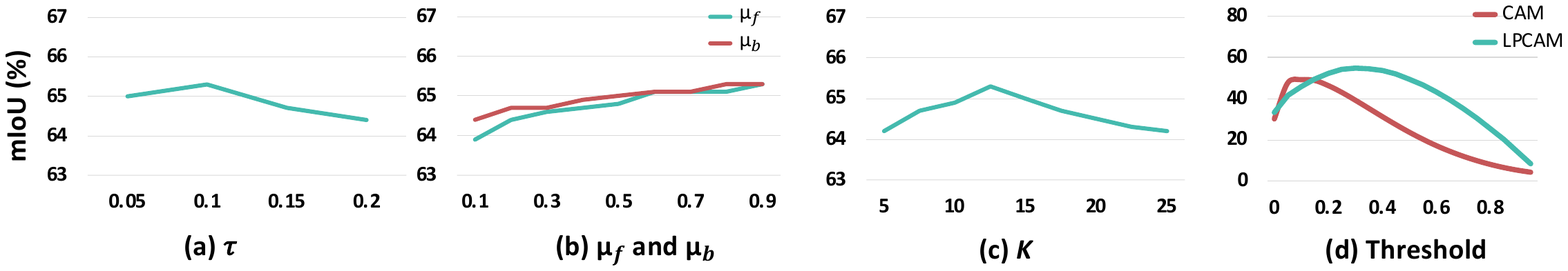}
    \vspace{-2mm}
    \caption{Sensitivity analysis on VOC, in terms of (a) $\tau$ for dividing foreground and background local features, (b) $\mu_f$ for selecting class prototypes and $\mu_b$ for selecting context prototypes, (c) the number of clusters $K$ in k-Means, and (d) the threshold used to generate 0-1 seed masks from heatmaps (a common hyperparameter in all CAM-based methods). \emph{Please refer to the supplementary materials for the results on MS~COCO.}}
    \vspace{-4mm}
    \label{fig_sensitivity}
\end{figure}
\setlength{\tabcolsep}{1.6mm}{
\renewcommand\arraystretch{1.1}
\begin{table}[ht]
  \centering
  \scalebox{0.9}{
  \begin{tabular}{llcccc}
    \toprule
    &\multirow{2}*{Methods} & \multirow{2}*{Sal.} &   \multicolumn{2}{c}{VOC} & MS~COCO \\
    \cmidrule(r){4-5}\cmidrule(r){6-6}
    &&&\texttt{val}&\texttt{test}&\texttt{val}\\
    \hline
    \multirow{13}*{\rotatebox{90}{ResNet-50}}
    &IRN~\cite{irn}          \tiny{CVPR'19}     &              & 63.5       & 64.8          & 42.0  \\
    &LayerCAM~\cite{layercam}\tiny{TIP'21}      &              & 63.0       & 64.5          & -     \\
    &AdvCAM~\cite{advcam}    \tiny{CVPR'21}     &              & 68.1       & 68.0          & 44.2  \\
    &RIB~\cite{rib}          \tiny{NeurIPS'21}  &              & 68.3       & 68.6          & 44.2  \\
    &ReCAM~\cite{recam}      \tiny{CVPR'22}     &              & 68.5       & 68.4          & 42.9  \\
    &\cellcolor{Gray}IRN+\texttt{LPCAM}    &\cellcolor{Gray} & \cellcolor{Gray}68.6    & \cellcolor{Gray}68.7      & \cellcolor{Gray}44.5  \\
    &SIPE~\cite{sipe}        \tiny{CVPR'22}     &              & 68.8       & 69.7          & 40.6  \\
    &OOD~\cite{ood}+Adv      \tiny{CVPR'22}     &              & 69.8       & 69.9          & -     \\
    &AMN~\cite{amn}          \tiny{CVPR'22}     &              & 69.5       & 69.6          & 44.7  \\
    &\cellcolor{Gray}AMN+\texttt{LPCAM}    &\cellcolor{Gray} & \cellcolor{Gray}70.1    &\cellcolor{Gray} 70.4      & \cellcolor{Gray}45.5  \\ 
    &ESOL~\cite{esol}        \tiny{NeurIPS'22}  &              & 69.9$^*$   & 69.3$^*$      & 42.6  \\
    &CLIMS~\cite{clims}      \tiny{CVPR'22}     &              & 70.4$^*$   & 70.0$^*$      & -     \\
    &EDAM~\cite{edam}        \tiny{CVPR'21}     &\checkmark    & 70.9$^*$   & 71.8$^*$      & -     \\
    &\cellcolor{Gray}EDAM+\texttt{LPCAM}  &\cellcolor{Gray}\checkmark & \cellcolor{Gray}71.8$^*$ &\cellcolor{Gray} 72.1$^*$& \cellcolor{Gray}42.1\\
    \hline
    \multirow{9}*{\rotatebox{90}{WideResNet-38}}
    &Spatial-BCE~\cite{sbce} \tiny{ECCV'22}     &              & 70.0       & 71.3      & 35.2  \\
    &BDM~\cite{bdm}          \tiny{ACMMM'22}    &\checkmark    & 71.0       & 71.0      & 36.7  \\ 
    &RCA~\cite{rca}+OOA      \tiny{CVPR'22}     &\checkmark    & 71.1       & 71.6      & 35.7  \\
    &RCA~\cite{rca}+EPS      \tiny{CVPR'22}     &\checkmark    & 72.2       & 72.8      & 36.8  \\
    &HGNN~\cite{hgnn}        \tiny{ACMMM'22}    &\checkmark         & 70.5$^*$   & 71.0$^*$  & 34.5  \\ 
    &EPS~\cite{eps}          \tiny{CVPR'21}     &\checkmark         & 70.9$^*$   & 70.8$^*$  & -     \\
    &RPIM~\cite{rpim}        \tiny{ACMMM'22}    &\checkmark         & 71.4$^*$   & 71.4$^*$  & -     \\ 
    &L2G~\cite{l2g}          \tiny{CVPR'22}     &\checkmark         & 72.1$^*$   & 71.7$^*$  & 44.2  \\
    \hline
    \multirow{2}*{\rotatebox{90}{\small{DeiT-S}}}
    &MCTformer~\cite{mctformer}    \tiny{CVPR'22}     &                 & 71.9$^{\dag}$  & 71.6$^{\dag}$   & 42.0  \\
    &\cellcolor{Gray}MCTformer+\texttt{LPCAM}      &\cellcolor{Gray} & \cellcolor{Gray}72.6$^{\dag}$  & \cellcolor{Gray}72.4$^{\dag}$  &\cellcolor{Gray} 42.8 \\
    \bottomrule
  \end{tabular}}
  \vspace{-2mm}
  \caption{The mIoU results (\%) based on DeepLabV2 on VOC and MS~COCO. The side column shows three backbones of multi-label classification model. ``Sal.'' denotes using saliency maps. * denotes the segmentation model is pre-trained on MS~COCO. $^\dag$ denotes the segmentation model is pre-trained on VOC.
  }
  \vspace{-6mm}
  \label{table_related}
\end{table}
}

\noindent
\textbf{Sensitivity Analysis for Hyperparameters.}
In Figure~\ref{fig_sensitivity}, we show the quality (mIoU) of generated seed masks when plugging LPCAM in AMN on VOC dataset. We perform hyperparameter sensitivity analyses by changing the values of (a) the threshold $\tau$ for dividing foreground and background local features, (b) the threshold $\mu_f$ for selecting class prototypes and the threshold $\mu_b$ for selecting context prototypes, (c) the number of clusters $K$ in K-Means, and (d) the threshold used to generate 0-1 seed mask (a common hyperparameter in all CAM-based methods). Figure~\ref{fig_sensitivity}(a) shows that the optimal value of $\tau$ is $0.1$. Adding a small change does not make any significant effect on the results, e.g., the drop is less than $1\%$ if increasing $\tau$ to $0.2$. Figure~\ref{fig_sensitivity}(b) shows that the optimal values of $\mu_f$ and $\mu_b$ are both $0.9$. The gentle curves show that LPCAM is little sensitive to $\mu_f$ and $\mu_b$. This is because classification models (trained in the first step of WSSS) often produce overconfident (sharp) predictions~\cite{confident}, i.e., output probabilities are often close to $0$ or $1$. It is easy to set thresholds ($\mu_f$ and $\mu_b$) on such sharp values. In Figure~\ref{fig_sensitivity}(c), the best mIoU of seed mask is $65.3\%$ when $K$=$12$, and it drops by only $0.7$ percentage points when $K$ goes up to $20$. In Figure~\ref{fig_sensitivity}(d), LPCAM shows much gentler slopes than CAM around their respective optimal points, indicating its lower sensitivity to the changes of this threshold.

\noindent
\textbf{Qualitative Results.}
Figure~\ref{fig_vis} shows qualitative examples where LPCAM leverages both discriminative and non-discriminative local features to generate heatmaps and 0-1 masks.
In the leftmost two examples, CAM focuses on only discriminative features, e.g., the ``head'' regions of ``teddy bear'' and ``dog'', while our LPCAM has better coverage on the non-discriminative feature, e.g., the ``leg'',  ``body'' and ``tail'' regions. In the ``surfboard'' example, the context prototype term $-\bm{BG}_n$ in Eq.~\ref{eq:LPCAM} helps to remove the context ``waves''. In the rightmost example, we show a failure case: LPCAM succeeds in capturing more object parts of ``train'' but unnecessarily covers more on the context ``railroad''. We think the reason is the strong co-occurrence of  ``train'' and ``railroad'' in the images of ``train''. 

\noindent
\textbf{Comparing to Related Works.}
We compare LPCAM with state-of-the-art methods in WSSS. As shown in Table~\ref{table_related}, on the common setting (ResNet-50 based classification model, and ResNet-101 based DeepLabV2 segmentation model pre-trained on ImageNet), our AMN+LPCAM achieves the state-of-the-art results on VOC ($70.1\%$ mIoU on \texttt{val} and $70.4\%$ on \texttt{test}). On the more challenging MS~COCO dataset, our AMN+LPCAM (ResNet-50 as backbone) outperforms the state-of-the-art result AMN and all related works based on WRN-38.
\section{Conclusions}
We pointed out that the crux behind the poor coverage of the conventional CAM is that the global classifier captures only the discriminative features of objects. We proposed a novel method dubbed LPCAM to leverage both discriminative and non-discriminative local prototypes to generate activation maps with complete coverage on objects. Our extensive experiments and studies on two popular WSSS benchmarks showed the superiority of LPCAM over CAM.

\noindent
\textbf{Acknowledgments}
The author gratefully acknowledges the support of the Lee Kong Chian (LKC) Fellowship fund awarded by Singapore Management University, and the A*STAR under its AME YIRG Grant (Project No.A20E6c0101).

{\small
\bibliographystyle{ieee_fullname}
\bibliography{main}
}

\clearpage
\beginsupp

\noindent
{\Large {\textbf{Supplementary materials}}}
\\

We present 
empirical validation about the classifier weights biased on discriminative dimensions in Section~\ref{sec:bias},
the impact of different clustering and similarity methods in Section~\ref{sec:similarity},
the analysis on the number of samples in clustering in Section~\ref{sec:num_samples},
ablation study on the MS~COCO dataset in Section~\ref{sec:ablation} supplementing for Table~\ref{table_ablation} (main paper), 
Sensitivity analysis on VOC in Section~\ref{sec:sensitivity} supplementing for Figure~\ref{fig_sensitivity} (main paper), and more qualitative results in Section~\ref{sec:qualitative} supplementing for Figure~\ref{fig_vis} (main paper).

\section{Biased Classifier}
\label{sec:bias}
\begin{figure}[ht]
\begin{center}
\includegraphics[width=0.46\textwidth]{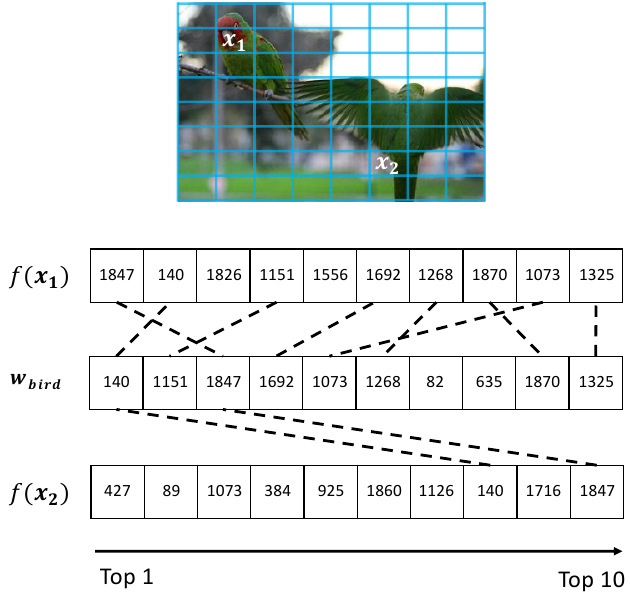}
\end{center}
  \caption{Empirical validation about the classifier weights biased on discriminative dimensions. We show the indices of the top 10 dimensions with the highest value for local region features ($f(\bm{x_1)}$: ``head'' and $f(\bm{x_2})$: ``tail'') and the classifier weight of ``bird'' ($\mathbf{w}_{bird}$).}
\label{fig:bias}
\end{figure}
We empirically validate that the classifier weights biased on the discriminative dimensions. This is to supplement for Section~\ref{sec_justfication} in the main paper.

In Figure~\ref{fig:bias} We show the indices of the top 10 dimensions with the highest value for local region features ($f(\bm{x_1)}$: ``head'' and $f(\bm{x_2})$: ``tail'') and the classifier weight of ``bird'' ($\mathbf{w}_{bird}$). The number of overlap dimensions between $\mathbf{w}_{bird}$ and $f(\bm{x_1)}$ is 8, but only 2 between $\mathbf{w}_{bird}$ and $f(\bm{x_2})$. This validate that the classifier weight of ``bird'' ($\mathbf{w}_{bird}$) biased on the dimensions of discriminative feature ``head''.

\section{Impact of clustering and similarity methods}
\label{sec:similarity}
We study the impact of different clustering and similarity methods.
1)~For clustering, we use K-Means and Hierarchical clustering. 
On VOC, the seed mask quality (mIoU) of Hierarchical clustering is 55.1\% (slightly higher than the 54.9\% of K-Means in Table~\ref{table_plugin}), but the running time is around 5 times longer.
2)~For similarity, we evaluate Euclidean and Cosine similarities in K-Means. The seed mask quality (mIoU) of using Euclidean on VOC dataset is 54.4\%, which is close to that of Cosine (54.9\% in Table~\ref{table_plugin}).

\section{Number of samples in clustering}
\label{sec:num_samples}
\begin{wrapfigure}{r}{4.5cm}
    \vspace{-1mm}
    \centering
    \includegraphics[width=0.9\linewidth]{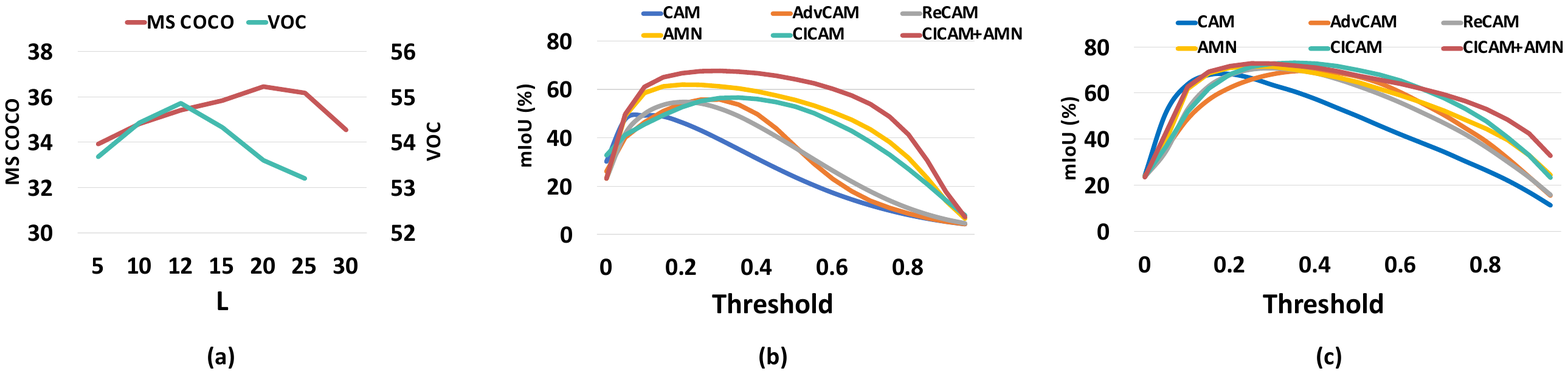}
    \caption{The seed mask quality (mIoU) of LPCAM regarding the number of images per class on MS~COCO.}
    \label{fig:samples}
    \vspace{-4mm}
\end{wrapfigure}
As mentioned in Section~\ref{sec_datasets} of the main paper: for k-Means clustering on MS~COCO, we sample $100$ images per class, rather than using the whole dataset of each class (to control the time costs for clustering). Here We study the impact of the number of images per class on the seed mask quality (mIoU) of LPCAM and show the results in Figure~\ref{fig:samples}. There is little performance gain after $100$. The possible reason is that for the common object on MS~COCO, $100$ samples can cover the variants of local features in the class.

\section{Ablation Study on MS~COCO}
\label{sec:ablation}

\setlength{\tabcolsep}{2.2mm}{
\renewcommand\arraystretch{1.1}
\begin{table}[ht]
    \centering
    \scalebox{0.8}{
      \begin{tabular}{lccccc}
        \toprule
                      & FP    & FN    & mIoU  & Prec. & Recall\\ 
        \midrule
        \texttt{CAM}          & 45.7                   & 21.9                   & 33.1                   & 43.8                   & 64.6  \\
        \texttt{LPCAM-F}      & 49.7\scriptsize{+4.0}  & 16.9\scriptsize{-5.0}  & 33.9\scriptsize{+0.8}  & 42.2\scriptsize{-1.6}  & 68.5\scriptsize{+3.9}  \\
        \texttt{LPCAM}        & 43.5\scriptsize{-2.2}  & 21.2\scriptsize{-0.7}  & 35.4\scriptsize{+2.3}  & 47.1\scriptsize{+3.3}  & 64.7\scriptsize{+0.1}  \\
        \bottomrule
      \end{tabular}
    }
    \caption{An ablation study on MS~COCO dataset. ``-F'' denotes only the ``Foreground'' term $\bm{FG}_n$ is used in Eq.~\ref{eq:LPCAM} (main paper).}
    \label{table_ablation_coco}
\end{table}
}

We conduct an ablation study on the MS~COCO dataset to evaluate the two terms of LPCAM in Eq.~\ref{eq:LPCAM} (main paper): foreground term $\bm{FG}_n$ and background term $\bm{BG}_n$ that accord to class and context prototypes, respectively. This is to supplement for Table~\ref{table_ablation} in the main paper. In Table~\ref{table_ablation_coco}, we show the mIoU results (of seed masks), false positive (FP), false negative (FN), precision, and recall. We can see that our method of using class prototypes (LPCAM-F) greatly improve the recalls---$3.9\%$ higher than CAM, and thus reduces the rates of FN a lot. This validates the ability of our methods to capture non-discriminative regions of the image. We also notice that LPCAM-F increases the rate of FP over CAM. The reason is that confusing context features (e.g., ``railroad'' for ``train'') may be wrongly taken as class features. Fortunately, when we explicitly resolve this issue by applying the negative context term $-\bm{BG}_n$ in LPCAM, this rate can be reduced (by $6.2\%$ for MS~COCO), and the overall performance (mIoU) can be improved (by $1.5\%$ for MS~COCO).

\begin{figure}[ht]
    \begin{center}
    \includegraphics[width=0.47\textwidth]{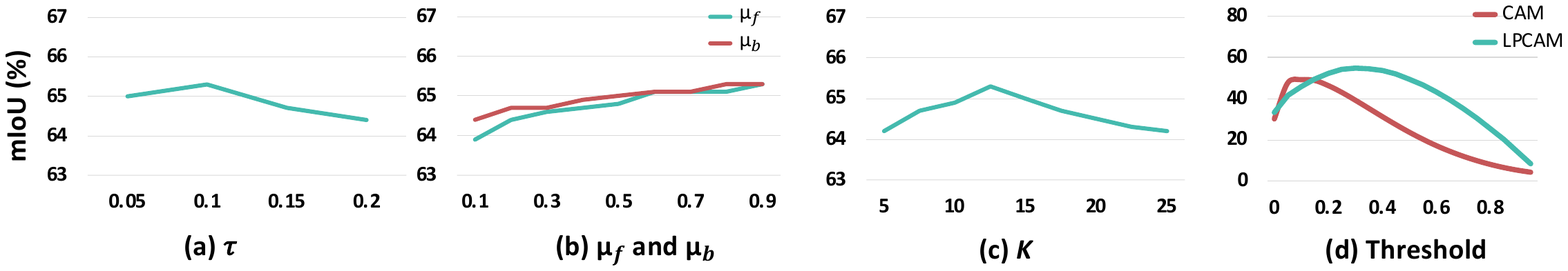}
    \end{center}
    \caption{Sensitivity analysis on MS~COCO, in terms of (a) $\tau$ for dividing foreground and background local features, (b) $\mu_f$ for selecting class prototypes and $\mu_b$ for selecting context prototypes, (c) the number of clusters $K$ in k-Means, and (d) the threshold used to generate 0-1 seed masks from heatmaps.}
    \label{fig:sensitivity_mscoco}
\end{figure}
\section{Sensitivity Analysis on MS~COCO}
\label{sec:sensitivity}
In Figure~\ref{fig:sensitivity_mscoco}, we show the quality (mIoU) of generated seed masks when plugging LPCAM in AMN on VOC dataset. We perform hyperparameter sensitivity analyses by changing the values of (a) the threshold $\tau$ for dividing foreground and background local features, (b) the threshold $\mu_f$ for selecting class prototypes and the threshold $\mu_b$ for selecting context prototypes, (c) the number of clusters $K$ in K-Means, and (d) the threshold used to generate 0-1 seed mask (a common hyperparameter in all CAM-based methods). Figure~\ref{fig:sensitivity_mscoco}(a) shows that the optimal value of $\tau$ is $0.25$. Adding a small change does not make any significant effect on the results, e.g., the drop is less than $1\%$ if decreasing $\tau$ to $0.15$. We use a higher $\tau$ on MS COCO because the quality of CAM on MS COCO is poorer (than VOC) and a higher value can filter out noisy activation. Figure~\ref{fig:sensitivity_mscoco}(b) shows that the optimal values of $\mu_f$ and $\mu_b$ are $0.9$ and $0.5$, respectively. The gentle curves show that LPCAM is little sensitive to $\mu_f$ and $\mu_b$. This is because classification models (trained in the first step of WSSS) often produce overconfident (sharp) predictions, i.e., output probabilities are often close to $0$ or $1$. It is easy to set thresholds ($\mu_f$ and $\mu_b$) on such sharp values. In Figure~\ref{fig:sensitivity_mscoco}(c), the best mIoU of seed mask is $42.5\%$ when $K$=$20$, and it drops by only $0.8$ percentage points when $K$ goes up to $30$. In Figure~\ref{fig:sensitivity_mscoco}(d), LPCAM shows much gentler slopes than CAM around their respective optimal points, indicating its lower sensitivity to the changes of this threshold.

\begin{figure*}[t]
    \includegraphics[width=0.98\textwidth]{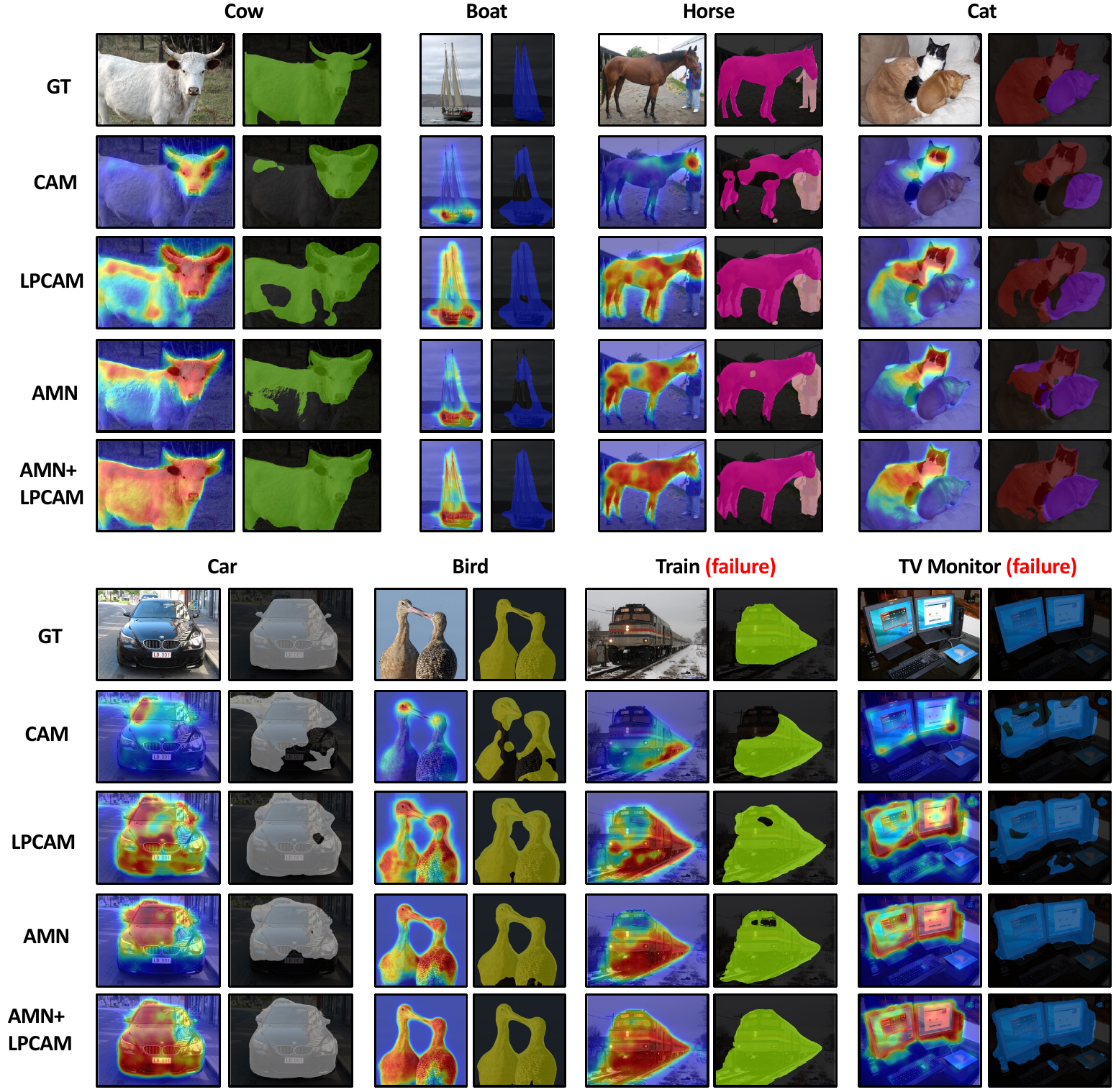}
    \caption{Qualitative results on VOC. In each example pair, the left is heatmap and the right is seed mask. }
    \label{fig:qualitative}
\end{figure*}
\section{Qualitative Results on VOC}
\label{sec:qualitative}
Figure~\ref{fig:qualitative} shows qualitative examples where LPCAM leverages both discriminative and non-discriminative local features to generate heatmaps and 0-1 masks on VOC dataset. In both single-object images (``cow'', ``boat'', ``car'', and ``bird'') and multi-objects images (``horse'' and ``cat''), CAM focuses on only discriminative features e.g., the ``head'' regions of ``cow'', while our LPCAM has better coverage on the non-discriminative feature, e.g., the ``body'' and ``leg'' regions. In the ``car'' example, the context prototype term $-\bm{BG}_n$ in Eq.~\ref{eq:LPCAM} (main paper) helps to remove the context ``plants''. In the last two examples, we show two failure cases: LPCAM succeeds in capturing more object parts of ``train'' and ``TV Monitor'' but unnecessarily covers more on the context ``railroad'' and ``keyboard''. We think the reason is the strong co-occurrence of  ``train'' and ``railroad'' in the images of ``train'' (``TV Monitor'' and ``keyboard'' in the image of ``TV Monitor''). 

\end{document}